\documentclass{article}
\usepackage[utf8]{inputenc}
\usepackage{authblk}
\usepackage{graphicx}
\usepackage[a4paper]{geometry}

\usepackage{changepage}
\usepackage{mathtools}
\usepackage{nameref,hyperref}
\usepackage{multirow}   
\usepackage[table,xcdraw]{xcolor}
\usepackage{todonotes}
\usepackage[]{algorithm2e}
\usepackage{acronym}
\usepackage{subcaption}
\usepackage{float} 

\RestyleAlgo{boxruled}
\SetAlgoSkip{bigskip}

\title{An Artificial Chemistry Implementation of a Gene Regulatory Network}
\author[1,2*]{Iliya Miralavy}
\author[1,2]{Wolfgang Banzhaf}
\affil{BEACON Center of Evolution in Action,}
\affil[2]{Department of Computer Science and Engineering}
\affil{Michigan State University, East Lansing, Michigan, United States}
\date{}                     
\affil[*]{Email: miralavy@msu.edu}
\begin{document}

\maketitle

\newacro{GRN}{Gene Regulatory Networks}
\newacro{ARN}{Artificial Regulatory Networks}
\newacro{CA}{Cellular Automata}
\newacro{SCA}{Stochastic Cellular Automata}
\newacro{AC}{Artificial Chemistry}
\newacro{BN}{Boolean Networks}
\newacro{PBN}{Probabilistic Boolean Networks}
\newacro{PN}{Petri Nets}
\newacro{ANN}{Artificial Neural Networks}
\newacro{GA}{Genetic Algorithm}
\newacro{TF}{Transcription Factors}

\begin{abstract}
Gene Regulatory Networks are networks of interactions in biological organisms responsible for determining the production levels of proteins and peptides. Proteins are workers of a cell factory, and their production defines the goal of a cell and its development. Various attempts have been made to model such networks both to understand these biological systems better and to use inspiration from understanding them to solve computational problems. In this work, a biologically more realistic model for gene regulatory networks is proposed, which incorporates Cellular Automata and Artificial Chemistry to model the interactions between regulatory proteins called the Transcription Factors and the regulatory sites of genes. The result of this work shows complex dynamics close to what can be observed in nature. Here, an analysis of the impact of the initial states of the system on the produced dynamics is performed, showing that such evolvable models can be directed towards producing desired protein dynamics. 
\end{abstract}

\section{Introduction}

Through millions of years, evolution has created and refined complex biological systems vital to the existence of natural organisms. Theoretically, a biological system is a network of interactions between natural entities that serve a specific purpose \cite{muggianu2018modeling}. For example, the lungs, trachea, nose, and related muscles work together to form the respiratory system in human beings responsible for breathing. Moreover, biological systems are not only limited to organic compound procedures consisting of different organs. Ant colonies are an excellent example of a system that is made of a population of complex living organisms serving the purpose of survival of the colony by many means, among them is spatial distribution of individuals in an ecosystem \cite{theraulaz2002spatial}. Most biological systems are adaptable, robust, and produce complex dynamics. Computational modeling of biological systems has been a topic of interest for many researchers due to the fascinating characteristics of these systems. Modeling can help infer meaningful and essential information in order to understand the complex underlying biological systems. Furthermore, it is possible to take inspiration from biological models to solve niche computationally represented problems and tasks. In this paper, we are particularly interested in modeling \ac{GRN}, which are complex networks of interactions between genes in a cell responsible for regulating metabolic flux through the production of enzymes. This work aims to introduce a more biologically realistic model of the \ac{GRN}s, investigate how the system's dynamics are affected by different initial states of the network, and discuss its possible applications.

\subsection{Gene Regulatory Networks}
\label{grnsubsec}

\begin{figure}[ht]
    \centering
    \includegraphics[width=0.7\columnwidth]{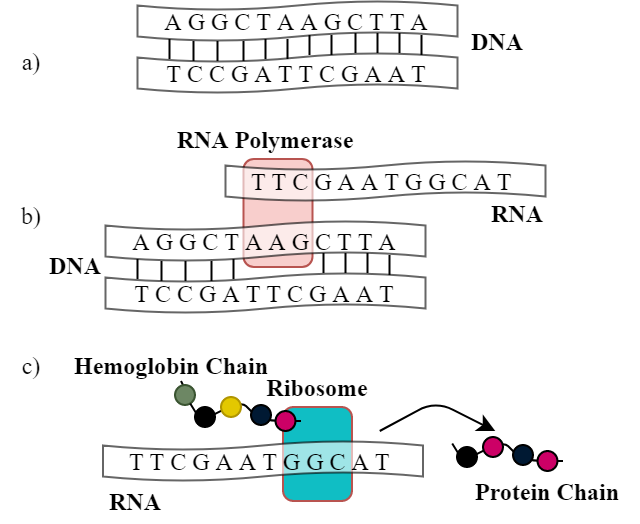}
    \caption{DNA to Protein. a) A double-stranded DNA. b) Transcription. DNA molecules move fast attached to an RNA polymerase which partially detaches the two strands to produce RNA. c) Translation. Ribosome translates codons in RNA sequence to their respective amino acids to create protein chains.}
    \label{fig:transcl}
\end{figure}

\begin{figure}[ht]
    \centering
    \includegraphics[width=0.7\columnwidth]{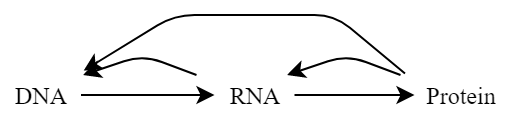}
    \caption{From DNA to Protein. A complex system of interactions}
    \label{fig:dnatoprot}
\end{figure}

Deoxyribonucleic acid, or DNA, is the genetic marker of all living organisms. DNA comprises two complementing strands of nucleotides connected with hydrogen bonds obeying a special base pairing rule. Nucleotides contain one of the four distinct bases that are commonly known as \textit{A} (Adenine), \textit{G} (Guanine), \textit{C} (Cytosine) and \textit{T} (Thymine). Because of their molecular structures, base \textit{A} only binds to \textit{T} and \textit{G} only binds to \textit{C} causing the two strands to become complementary. Exons are the expressed parts of the DNA molecules that code for proteins, while most DNA consists of non-coding regions known as introns. DNA is divided into genes which are basic heredity units of an organism and are passed along from parent to offspring \cite{Carroll2005}. Different nucleotides in a gene sequence determine the types of proteins produced. The generation of proteins from double-stranded DNA strings follows two main steps: Transcription and Translation. In transcription, this double-stranded DNA is divided into single DNA strands and then transcribed into RNA molecules, and in translation, RNA is converted into proteins via a coding mechanism. There are several regulatory regions on genes that control the production of proteins. For example, the promoter region determines the starting point of a gene to be transcribed. During transcription, RNA polymerase, a complex structure composed of protein sub-units, binds to the promoter region of a gene, separates the two DNA strands and replicates one of them to create RNA molecules (figure \ref{fig:transcl}b). In translation, cellular structures made of proteins and RNA called ribosomes use RNA codons (sets of three nucleotides) as templates for creating sequences of amino acids (figure \ref{fig:transcl}c). A complete sequence of amino acid, forms a protein \cite{watson2009}\cite{calladine2004}. Enhancer and inhibitor regions are regulatory sites located upstream or downstream of the promoter region of a gene \cite{pennacchio2013enhancers}. A special class of regulatory proteins called \ac{TF} can bind to the enhancer region of a gene to increase the likelihood of its transcription. In contrast, binding to the inhibitor region would repress the transcription rate of that particular gene. Figure \ref{fig:dnatoprot} shows a high-level overview of the process of producing proteins from DNA. RNA molecules as intermediate products and proteins as end products of this process also serve as the regulators in this system, creating feedback loops and forming a network of interactions between genes. This complex network of interactions that control the cell production is called the Gene Regulatory Networks \cite{Levine2005}. 

GRNs differentiate between cells to form different biological tissues, control cell metabolism, help with cell signal transduction and determine the body shape and behavior of complex organisms. Unraveling these networks' mysteries is especially important for better understanding the DNA life cycle and can have applications such as identifying and curing genetic disorders \cite{Gnanakkumaar2019}. Modeling \ac{GRN} dynamics in computational frameworks has been a direct approach to studying these networks. Also, the computational representations of \ac{GRN}s have been applied to solve various computational problems. However, due to the complex nature of \ac{GRN}s, these models are often mathematical abstractions of their biological counterparts and do not account for the stochastic nature of their building blocks\cite{Arias2014}, which is a shortcoming of such models \cite{ruth1997modeling}. 

\subsection{Modeling Biological Systems}

There has been substantial progress in modeling biological and physical systems throughout the literature \cite{deakin1990modelling, wilkinson2009stochastic, villaverde2019computational}. In general, it is possible to classify biological models into three types: \textit{Static}, \textit{Comparative Static} and \textit{Dynamic} \cite{ruth1997modeling}. \textit{Static} models are a snapshot of a process or an event—for example, a map of pandemic intensity in different regions of the world at a certain time.  \textit{Comparative Static} models are several snapshots of a process or event at different times that can be used to compare and retrieve meaningful data without modeling the process itself. Finally, \textit{Dynamic} models aim to model a sequence of processes and events by representing the changes in the state of a system over time. For example, a differential equation showing and predicting the spread of a pandemic over time is a dynamic model \cite{mohamadou2020review}. Dynamic models are not limited to mathematical approaches. Depending on the system, many approaches such as discrete-event modeling \cite{wainer2017discrete} or agent-based modeling \cite{holcombe2012modelling} are becoming important. 

In nature, \ac{GRN}s are pretty complex, with numerous elements playing different roles for the processes to work (some of which are described in Section \ref{grnsubsec}). Accounting for every detail of these systems is quite a rigorous task and might not be the best approach in a computational framework. First, not all the aspects of \ac{GRN}s are discovered or fully understood as of yet. Second, including every known detail of these systems in a biologically-perfect model would require massive computational resources and drastically limits their applications. Moreover third, previous literature has shown that abstract models can still give a good approximation of the dynamics and the processes of their natural counterparts. Models are always simplifications of a system, and it is a question of what function the models should serve, and which simplifications are justified. We base our work on \ac{ARN} introduced in \cite{Banzhaf2003}, a model inspired by natural gene regulation that uses interactions between genomes represented by bit strings to form \ac{GRN} networks. Two approaches of \ac{AC} and \ac{CA} are used to build the proposed \ac{GRN} model. \ac{AC}s bring aspects of agent-based dynamics and \ac{CA}s bring spatial aspects to the system in abstract forms. 


\subsection{Artificial Chemistry}

\ac{AC} is a sub-field of Artificial Life and, in computational frameworks, can be described as an artificial chemical system similar to a natural chemical system. An \ac{AC} makes it possible to model molecular interactions by defining rules for binding and detachment of artificial molecules, which can form new entities or regulate different processes. More formally, an \ac{AC} can be denoted as a triple\textit{ (S, R, A)} in which \textit{S} is a set of available molecules, \textit{R} is a set of all possible interaction rules, and \textit{A} is an algorithm that describes the system and how the molecules or objects interact with each other \cite{Dittrich2001}. In a case in which molecules can move, an \ac{AC} allows for rich and more complex interactions to emerge in the system \cite{hutton2002evolvable}, which is in line with the goals of this paper. \ac{AC}s have been previously used to model neural networks \cite{astor2000developmental}, self-organizing systems \cite{dittrich1998mesoscopic} and self-replicating systems \cite{kruszewski2022emergence}.  

\subsection{Cellular Automata}
A \ac{CA} is a discrete temporospatial model that can be described as a lattice network of cells that can have $N \geq 2$ states. Generally, each cell has a quiescence state that can change over time following the \ac{CA} rules. In most cases, these rules apply to all the cells of the grid and do not change over time; however, in the case of \ac{SCA}, cell states change depending on probabilities of a random distribution \cite{schiff2011cellular}. Changes in cell states over time caused by internal system rules can create exciting phenomena. 

In this work, the \ac{ARN} originally proposed in \cite{Banzhaf2003} is combined with an \ac{AC} in conjunction with a \ac{SCA}. The proposed system is less abstract than the original representation and generates various complex protein concentration dynamics while modeling complex DNA-protein interactions. The rest of the paper is summarized as follows: Section \ref{relatedsec} explores the previous literature for modeling \ac{GRN}s. Then, Section \ref{methodsec} explains the methods and algorithms used to define the proposed model. Next, Section \ref{resultsec} shows the results for the dynamics produced by the proposed system based on different initial conditions. Finally, Section \ref{discussionsec} discusses our results and points out the possible future directions of the current research.

\section{Related Literature}
\label{relatedsec}

\ac{GRN}s have been modeled using various principles in the literature. However, logical and discrete models are the most straightforward approaches to model \ac{GRN}s \cite{Karlebach2008}. In these methods, GRNs are considered to have discrete states and time steps. In each time step, the system updates according to regulatory functions which might result in a change of state. 

\ac{BN} \cite{Glass_1973} and \ac{PBN} \cite{Shmulevich_2002} are the most common logical techniques to model GRNs. In \ac{BN}s, each gene has only two states, \textit{expressed} and \textit{not expressed}. The state of each gene in the current time step is determined by the state of other genes in the previous time step and the regulatory functions \ac{PBN} is a subset of \ac{BN} that accounts for the stochasticity in dynamic systems and gives insights into the biological GRNs. A substantial increase in the number of states in \ac{BN}s and \ac{PBN}s makes analyzing such systems difficult. 

\ac{PN} is a non-deterministic mathematical modeling approach that has been previously used to represent \ac{GRN}s \cite{Chaouiya_2011}. \ac{PN}s are made of \textit{transitions}, \textit{places} and \textit{arcs}. In each \textit{place} there can be zero or more \textit{tokens}. An \textit{arc} is an entity that connects a transition to a place or vice versa and has a weight. In a \ac{PN} a transition is enabled, if there are sufficient \textit{tokens} in the starting state, i.e a number equal or greater than the \textit{arc} weight. A \textit{transition} of a \ac{PN} may fire if it is enabled, in which case it consumes the \textit{tokens} in the starting state and creates \textit{tokens} in the output state. In biological modeling of \ac{GRN}s using \ac{PN}s, \textit{places} represent molecules, \textit{transitions} represent reaction rules and \textit{tokens} represent concentration levels \cite{cussat2019artificial}. Stochasticity of \ac{PN}s occurs when multiple transitions are enabled to the same place. In such cases, transitions may fire in any order. This uncertainty makes \ac{PN}s similar to \ac{PBN}s.

Fractal \ac{GRN}s are models that use \textit{fractal proteins} and pattern matching interaction rules to represent \ac{GRN}s. \textit{Fractal proteins} are a subset of the Mandelbrot set (a set of complex numbers) that can exist in an environment or an artificial cell. Apart from \textit{cell fractal proteins}, a cell contains cytoplasm, genome, and some behaviors. A receptor gene in the cell works like a mask that allows for specific protein patterns to enter the cell area. Proteins interact through their fractal shape and the genetic markers of the genome's regulatory sites to form a network of interaction \cite{bentley2003evolving}.       

An example of Artificial Chemistry has been previously used by Astor and Adami \cite{astor2000developmental} to model a regulatory network used for the evolution of \ac{ANN}. These authors used a hexagonal grid in which each cell could have a concentration of substrates produced by neurons. These substrates can be different types of proteins or neurotransmitters. In their system, proteins diffuse based on differential equations, and genes are expressed if there are enough chemicals of certain types in the cell's cytoplasm. The hexagonal grid they incorporated for their work could be characterized as a \ac{CA}. \ac{CA}s have been an excellent framework to model \ac{GRN} algorithms in other works as well. For example, \cite{chavoya2008cell} uses a \ac{GA} to evolve an \ac{ARN}s to solve the French Flag problem on a cellular automaton grid. 

\ac{GRN} models have been previously incorporated in various classes of applications. Some work focus on the dynamic analysis of such systems. \cite{cussat2012using} analyzes the complex patterns generated by the dynamics of \ac{ARN}s by generating pictures and videos from the change of the concentrations of proteins. They evolve \ac{ARN}s to produce more interesting patterns by asking human users to rate the fitness of the produced images. \cite{bentley2004evolving, joachimczak2009evolution, bongard2003evolving} used \ac{GRN} models to perform morphogenesis. An interesting characteristic of using \ac{GRN} models for this purpose is the emergence of repetitive patterns in the evolved shapes rather than chaotic ones. \ac{GRN} models have also been used in applications such as agent or robot control showing comparable performance with other AI methods \cite{asr2013new, sanchez2014gene}. Finally, indirect encoding has been a topic of interest for applying \ac{GRN} models \cite{wrobel2012evolving, wrobel2014using}. The compact and evolvable representation of \ac{GRN}s can produce massive networks of interactions which makes them a good candidate to indirectly code for other systems such as \ac{ANN}s


\section{Methodology}
\label{methodsec}

The model proposed accounts for the protein-gene interactions in a single artificial cell to produce protein concentration dynamics. This model represents \ac{GRN}s by a linear DNA sequence. The DNA sequence can have a number of genes that are identified by promoter and terminator regions in the sequence. Each gene codes for a specific type of protein that serves as a regulatory agent (\ac{TF}) to control the transcription rate of other genes, by binding to their regulatory sites. These regulatory sites are Enhancer and Inhibitor regions located downstream or upstream of a promoter sequence. Biologists determine the location of these regulatory sites by genome-wide location analysis \cite{jin2011enhancers}. Since this technique is not applicable to the proposed model, we determine the location of these regulatory sites on the artificial genome by applying an arbitrary rule to a designated region right after the promoter sequence.

\begin{figure}[ht] 
  \begin{subfigure}[b]{0.32\linewidth}
    \centering
    \includegraphics[width=0.95\linewidth]{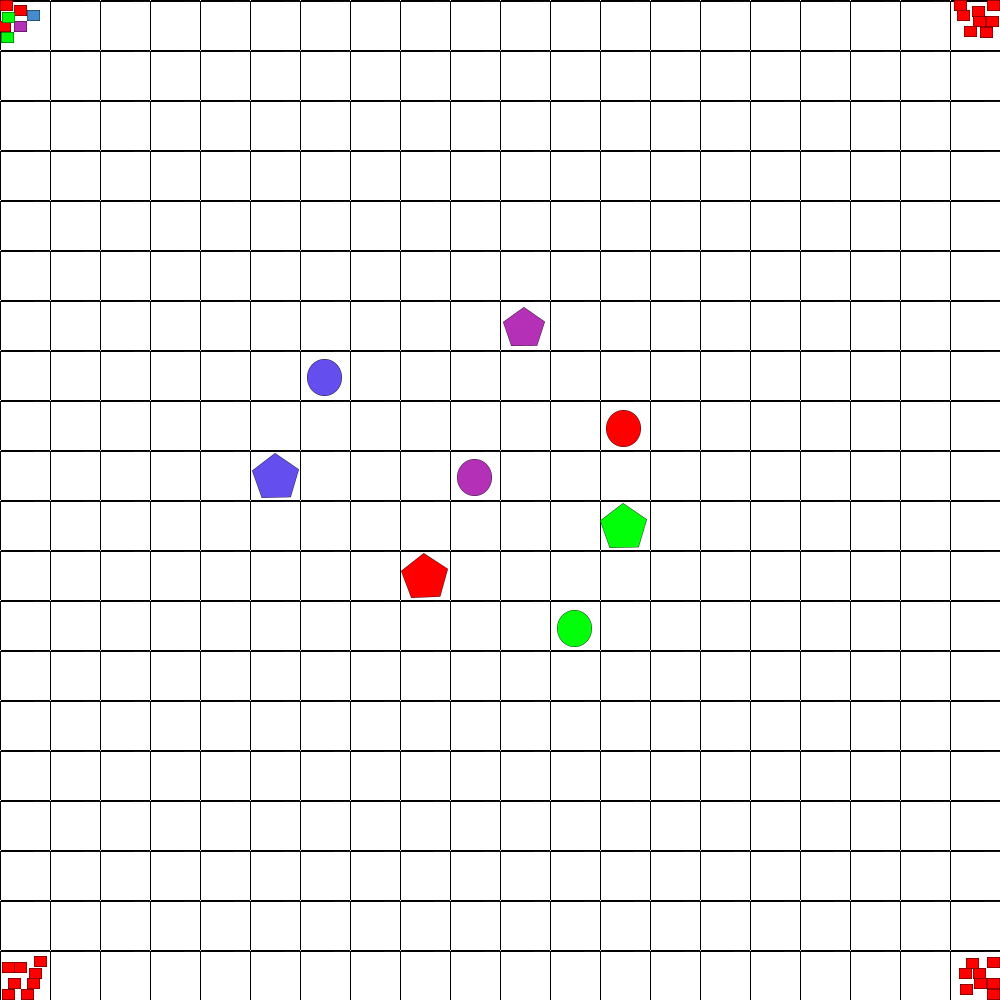} 
    \caption{Initial Stage} 
    \label{model:a} 
  \end{subfigure}
  \begin{subfigure}[b]{0.32\linewidth}
    \centering
    \includegraphics[width=0.95\linewidth]{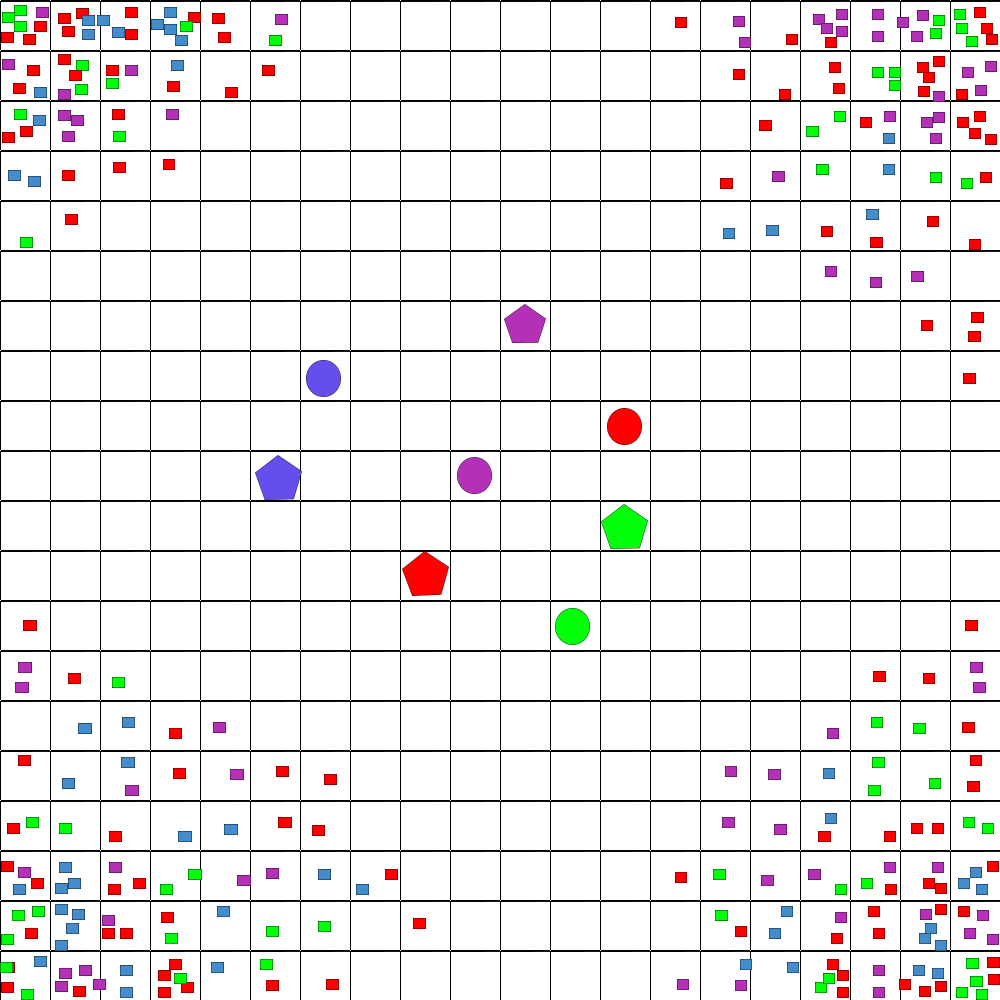} 
    \caption{Early Stage} 
    \label{model:b} 
  \end{subfigure} 
  \begin{subfigure}[b]{0.32\linewidth}
    \centering
    \includegraphics[width=0.95\linewidth]{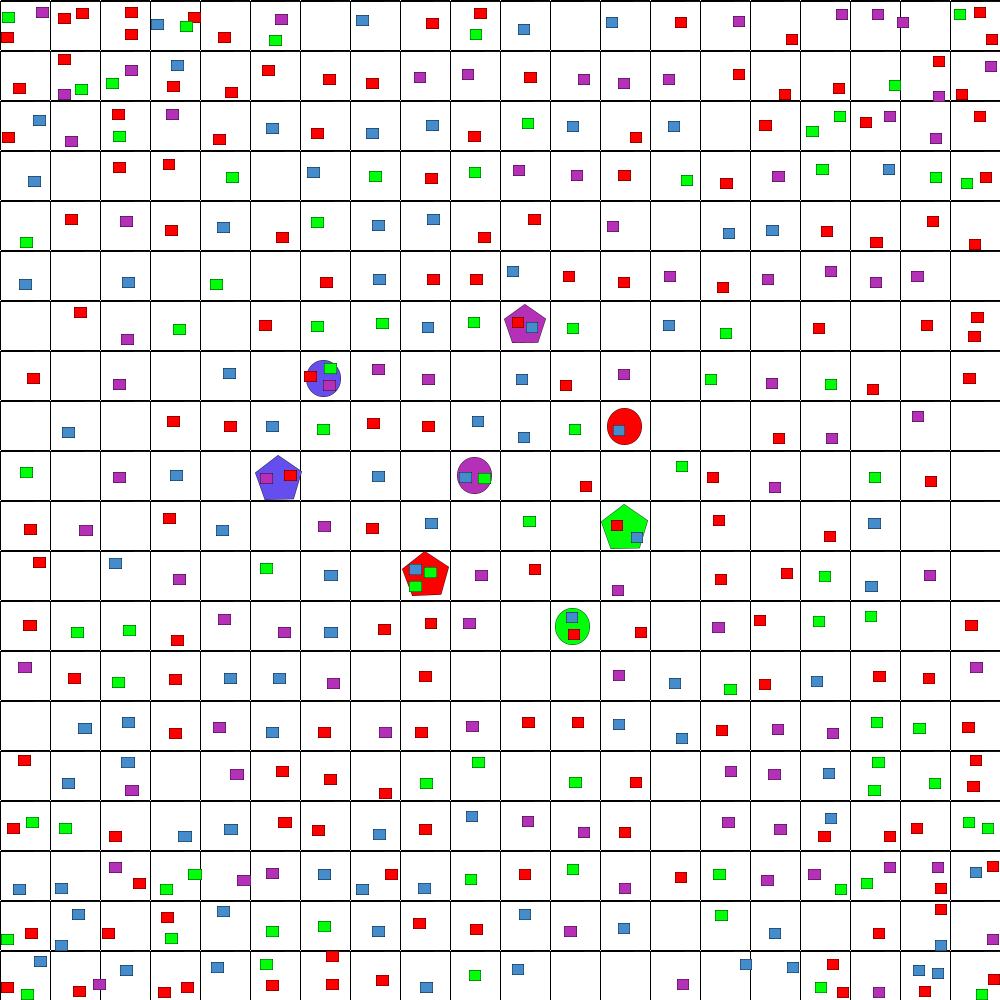} 
    \caption{Later Stage} 
    \label{model:c} 
  \end{subfigure}
  \caption{A snapshot of the 2D grid of the model showing four different genes and their \ac{TF}s in three different simulation stages. Different colors code for different genes. Pentagons represent the enhancer region of a gene, and circles represent the inhibitor sites. Small squares illustrate the different \ac{TF}s that move around the grid and can bind to the regulatory sites of genes other than their parent gene.}
  
  \label{fig_model} 
\end{figure}

When the genes and their regulatory sites are identified, they are placed randomly close to the center of a 2D grid representing the cell. An equal number of \ac{TF}s for each gene is initially positioned in a corner of the cell grid. In each regulatory time step of the system, these transcription factors can randomly move on the grid. If the location of a \ac{TF} is the same as a regulatory site of a gene, it may bind to that site. For example, binding to the enhancer site increases the transcription rate of that gene, and binding to the inhibitor site of a gene decreases that rate. \ac{TF}s stay bound for a certain number of regulatory cycles depending on the initial binding strength and, after that, are removed from the cell. When a \ac{TF} protein is removed, it will be replaced by a \ac{TF} belonging to a gene with the most protein concentration on that cycle. In the proposed system, a \ac{TF} cannot bind to the regulatory sites of its parent gene. Figure \ref{fig_model} illustrates three snapshots of this system in three stages of the simulation. 

The proposed model's \ac{AC} is configured by applying the frequently used techniques introduced in \cite{Banzhaf2015}.

\textbf{2D Space:} Spatial properties play an important role in the biological factories of a cell. \ac{CA} approaches can introduce spatial properties to a system and are incorporated in this work. A 2D grid is used to represent the biological cell. This approach enables us to measure the distance between entities and allow them to move around the grid while not being too computationally expensive. Entities might overlap on the same grid space. Grid borders are interconnected, meaning that if an entity moves out of one side, it will come back to the grid from the side of the opposing border, continuing the move in the same direction. 

\textbf{Time Measurement:} We use the notion of \textit{cycle} to determine regulatory time steps in the proposed system. In each cycle, the \ac{ARN} goes through a movement phase in which system entities perform a limited random walk in the 2D grid. This process is followed by a regulation phase in which the spatial development in the previous phase enables nearby entities to interact with one another and provides the basis for change in the system dynamics.

\textbf{Pattern Matching:} We use pattern matching as the main rule of interaction between the different entities in the system. This technique is not too different from its biological counterparts, where nucleotide bases' shape allows hydrogen bonds to occur between pairs. 

\subsection{Set of System Entities (\textit{S})} 

The artificial entities in \textit{S} are defined as follows:

\textbf{DNA:} DNA molecules are modeled as a linear sequence of bases (\textit{A, G, C} and \textit{T}). DNA is randomly initialized at the very start, determines the structure of the network of interactions, and consists of a number of genes and introns. In biology, DNA is made of two complementary strands; however, here we simplify it to modeling only one strand. This sequence is not modeled spatially in the \ac{CA} and only serves as the genome representation of individual \ac{GRN}s.

\begin{figure}[ht]
    \centering
    \includegraphics[width=0.9\columnwidth]{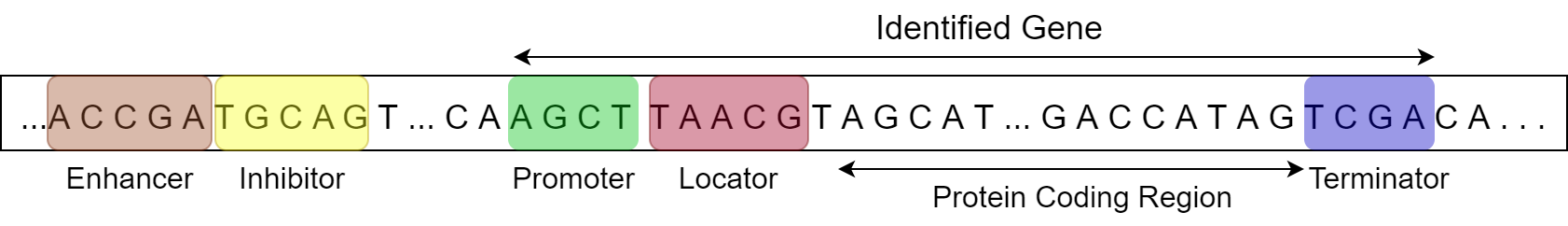}
    \caption{An identified gene in a DNA sequence. The promoter sequence (green) determines the gene's starting point, and the terminator sequence (purple) specifies the endpoint. Genes have enhancers and inhibitors regulatory sites that can be downstream or upstream of the promoter sequence (brown and yellow). The locator site determines the location of the inhibitor and enhancer sites.}
    \label{fig:gene}
\end{figure}

\textbf{Gene:} A gene is a subset of DNA that starts and ends with unique patterns of bases. Arbitrary four-base patterns of \textit{"AGCT"} and \textit{"TCGA"} are chosen to determine the start and the end of a gene, respectively. The starting pattern of a gene resembles the promoter region in biological genes. Gene identification happens in the system after the DNA is initialized and might result in identifying genes with different lengths. Genes code for proteins and have two regulatory sites of \textit{enhancer} and \textit{inhibitor} regions that regulate the protein production in the artificial cell and are located upstream or downstream of the promoter region of the DNA. The position of these regulatory sites is coded in the genome and is determined by following a simple rule. The length of these regions depends on the length of the gene. Genes with a more extensive sequence have larger regulatory sites and vice versa. The sequence of bases between the promoter and the end pattern of a gene determines the genetic marker of the proteins produced by that gene and the position of the enhancer and the inhibitor sites (Figure \ref{fig:gene}). 

\begin{figure}[ht]
    \centering
    \includegraphics[width=0.8\columnwidth]{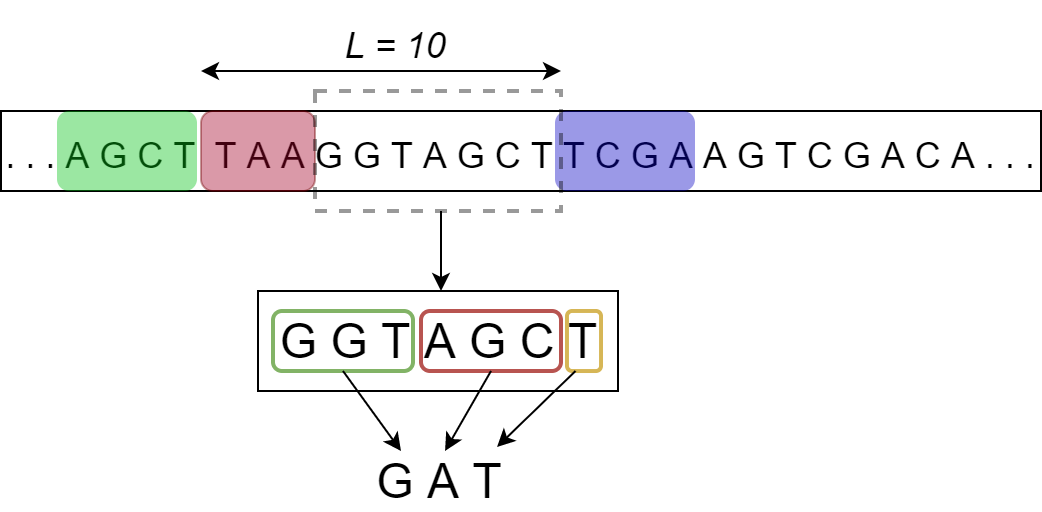}
    \caption{An example for calculating the protein sequence of a gene. It is allowed for the last chunk to have a lower number of bases.}
    \label{fig:protein}
\end{figure}

The size of the regulatory sites, including the inhibitor, enhancer, locator, and the coded protein, are the same and are calculated using the following equation:

$$S = \lfloor \sqrt{L} \rfloor$$

where $S$ denotes the size of the regulatory sites, and $L$ is the length of the sequence between, after the promoter and before the terminator sequence. An integer mapping is performed following an arbitrary rule to locate the location of the enhancer and the inhibitor regions. In this mapping, bases $T$, $G$, $C$ and $A$ are mapped to the values of $-1$, $-2$, $1$ and $2$, respectively. The Sum of the mapped values of the locator sequence determines the distance of the enhancer site (the inhibitor site is right next to the enhancer) from the gene's promoter sequence. For example, in Figure \ref{fig:protein}, the locator site has a sequence of $TAA$, and therefore, the enhancer region of the gene starts at $3$ distance from the promoter gene upstream of the DNA overlapping with the protein-coding region of the gene.

\textbf{Proteins:} Proteins are the end products of genes. Genes with higher transcription rates have higher produced protein concentrations. Each gene codes for a specific protein sequence. \ac{TF} proteins are modeled in the proposed system and are responsible for regulatory actions. The number of available TFs for each gene is relative to the protein production levels of that gene. In each cycle, TFs randomly walk in a cell and bind to the regulatory sites of genes if the distance between the two is less than a threshold value. As the system updates, the concentration levels of proteins vary based on the network of interactions between genes causing interesting dynamics to appear in the system. 

As mentioned above, the protein-coding region of the gene determines the genetic marker of the produced proteins. These proteins have the same length as the other regulatory sites, and their code is calculated following a majority rule. Figure \ref{fig:protein} illustrates how the protein sequence is determined. The length of the region between the promoter and the terminator sequences is $L = 10$ and therefore the size of the regulatory regions is $S = \lfloor \sqrt{10} \rfloor = 3$ and the locator site can be located (shown with the pink box in Figure \ref{fig:protein}). The protein coding region (surrounded by a dashed rectangle in Figure \ref{fig:protein}) has a size of 7. First, this region is divided into $S$ chunks with the size of $\leq N$ using the following formula:

$$ N = \lceil L / S \rceil $$

and then the majority rule applies to each chunk in a way that in each chunk the nucleotide with the higher frequency of occurrence gets selected as a base in the protein sequence. In the case of equal frequencies, the base with the first occurrence in the chunk gets selected.

\subsection{Set of System Rules of Interaction (\textit{R})} 

\begin{figure}[ht]
    \centering
    \includegraphics[width=1\columnwidth]{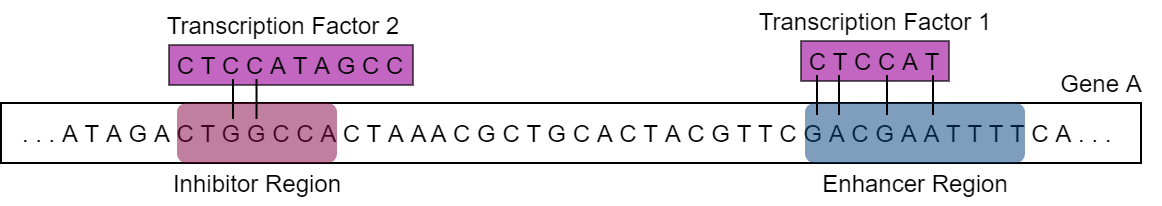}
    \caption{Binding between \ac{TF} and regulatory sites of two genes. These artificial bindings occur similarly to DNA nucleotide hydrogen bindings. The number of base-base bindings determines the binding strength. On the right side, Transcription Factor 1 is connected to the Enhancer region of gene $A$ with a binding strength of 4 that lasts for 4 cycles. On the left side, Transcription Factor 2 is connected to the Inhibitor region of gene B with a binding strength of 2 that lasts for 2 cycles.}
    \label{fig:tfbinding}
\end{figure}

Entirely modeling the Transcription and the Translation process with a computational viewpoint seems unnecessary. At each cycle of the regulation, first, the transcription rates of genes update with regards to the number of \ac{TF} proteins bound to that gene's regulatory sites. The next step is the moving phase, where all the \ac{TF}s can move around the artificial cell with a random walk approach to bind to the regulatory sites. Next, the \ac{TF}s within the binding range of the regulatory sites might bind to those sites with a binding strength determined by a nature-inspired binding rule. The binding strength is calculated by counting the number of base-base bindings of the regulatory site sequence and \ac{TF} sequence. Similar to nature, base \textit{A} only binds to \textit{T} and base \textit{G} only binds to \textit{C}. If the two sequences do not have the same length, the extra bases of the longer sequence are neglected. If the binding strength is zero, meaning that no \textit{A-T}, \textit{T-A}, \textit{G-C} or \textit{C-G} base-base binding could be found, then the binding does not happen. The binding strength indicates how many cycles the bounded \ac{TF} alters the transcription rate of the gene until the binding expires. Figure \ref{fig:tfbinding} illustrates the \ac{AC} binding method used in the proposed system. Also, \ac{TF}s cannot bind to their parent gene. Each artificial gene produces proteins at a minimal rate if no binding happens. However, if any binding happens, depending on which site (enhancer or inhibitor) the TF is binding to, the transcription rate or the concentration of proteins produced by that gene might alternate in that cycle.
 
The impact of the \textit{TF-Enhancer} and \textit{TF-Inhibitor} bindings on the translation rate of the respective gene is calculated using the following formulas:

$$ R_{i, t+1} = R_{i, t} + \frac{1}{N} \sum_{j=1}^N e^{\beta \times (S_{i,j} - S_{total} - 1)} $$
$$(TF-Enhancer)$$
$$ R_{i, t+1} = R_{i, t} - \frac{1}{N} \sum_{j=1}^N e^{\beta \times (S_{i,j} - S_{total} - 1)} $$
$$(TF-Inhibitor)$$

where $R_{i, t}$ refers to the transcription rate of gene $i$ at cycle $t$, $N$ is the total number of bindings to gene $i$, $\beta$ is an arbitrary constant and $S_{i,j}$ and $S_{total}$ are the binding strength between regulatory site of gene $i$ and TF $j$ and the strongest binding strength witnessed in the cycle, respectively.

Each gene produces a unique protein. At the end of the regulation cycle, protein concentrations update with the following formula:

$$C_{i,t+1} = C_{i, t} + \delta \times C_i \times R_i$$

where $C_{i, t}$ denotes concentration of protein $i$ at cycle $t$, $\delta$ is an arbitrary constant and $C_{bound}$ is the protein concentration level of the gene. After calculating the new protein concentrations, these values for each gene are normalized by dividing them by the total concentration of all proteins to keep the sum of the concentration levels equal to $1$ at all times. This approach will cause competition between concentration levels causing interesting dynamics to appear in the system. 

\begin{algorithm}[H]
\SetKwComment{Comment}{\# }{ }
\label{alg1}
 \KwData{TF\_list, gene\_list, binding\_threshold}
 \KwResult{Artificial Gene Regulatory Network}
 initialize\_grid() \\
 cycle $\gets$ 0 \\
 \While{cycle $<$ max cycle}{
    \For {TF in TF\_list} {
        \Comment{Update Transcription Rates}
        \If {TF.binding\_strength $>$ 0}{
            update\_rate(TF.bound\_gene) \\
            TF.binding\_strength $\gets$ TF.binding\_strength - 1 \\
            \If {$TF.binding\_strength == 0$}{
                remove\_tf(TF) \\
                create\_tf() \\
            }
            continue; \\
        }
        \Comment{Movement Phase}
        random\_walk(TF) \\
        \Comment{Binding Phase} 
        \For {gene in gene\_list} {
            \If {distance(TF, gene) $<$ binding\_threshold} {
                bind(TF, gene) \\    
            }
        }
    }
    \Comment{Production/Translation Phase}
    \For {gene in gene\_list} {
        update\_concentration(gene) \\
    }
    $cycle \gets cycle + 1$\;
 }
 \caption{Algorithm of the proposed \ac{GRN} model}
\end{algorithm}

\subsection{The Algorithm (A)}

Algorithm \ref{alg1} summarizes the algorithm used to model the \ac{GRN}s. Before the regulatory cycles start, the grid is initialized, and all the entities' positions are determined. Afterward, in each cycle, if a \ac{TF} is bound (binding strength $>$ 0), the transcription rates of the bound gene are updated depending on which regulatory site of the gene the \ac{TF} is connected to. If a \ac{TF} is not bound, it randomly walks around the 2D grid during the movement phase. Then the distance between every \ac{TF} and regulatory sites of every gene is checked, and if it is less than a specified binding threshold, the binding between the two entities happens. Finally, each gene's protein concentration is determined and normalized.

\section{Results}
\label{resultsec}
A series of experiments are conducted using the proposed \ac{ARN} model to show the varying dynamics produced by such systems as well as study how the initial state of the system affects the produced dynamics and how these dynamics can be evolved. 

\subsection{Varying Dynamics}

The proposed system produces various protein concentration dynamics. It is difficult to systematically quantify the produced dynamics regarding their complexity, similarity, or stability. Nonetheless, an attempt has been made to handpick and introduce a few of the outstanding patterns observed during our experiments. For all of these dynamics, a genome size of 3000 bases, 25 initial \ac{TF} proteins for each gene, a grid/cell size of 10, random walk step size of 5 for \ac{TF}s, $\beta$ and $\delta$ equal to $1$, and an initial concentration of $\frac{1}{N}$ was selected in which $N$ is the number of genes. The only factor that differentiates between the produced dynamics is random genomes. All the experiments were run for 1000 regulatory cycles. 

\begin{figure}[ht] 
  \begin{subfigure}[b]{0.45\linewidth}
    \centering
    \includegraphics[width=0.95\linewidth]{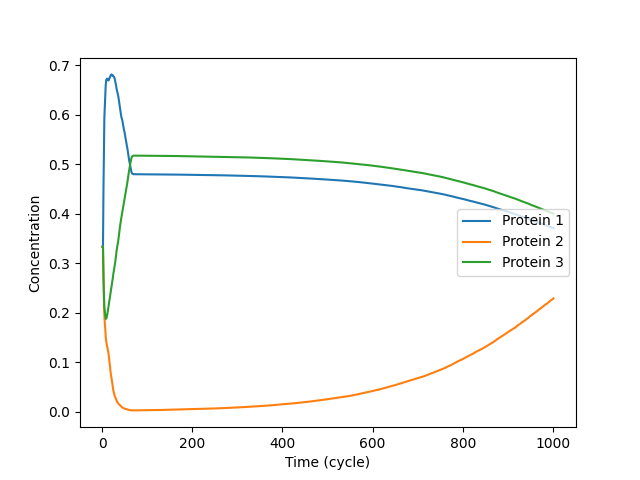} 
    \caption{Simple Development} 
    \label{dyna:a} 
  \end{subfigure}
  \begin{subfigure}[b]{0.45\linewidth}
    \centering
    \includegraphics[width=0.95\linewidth]{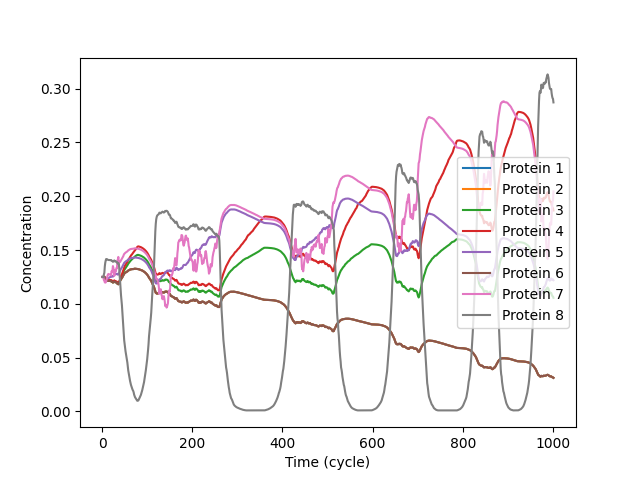} 
    \caption{Oscillatory} 
    \label{dyna:b} 
  \end{subfigure} 
  \begin{subfigure}[b]{0.45\linewidth}
    \centering
    \includegraphics[width=0.95\linewidth]{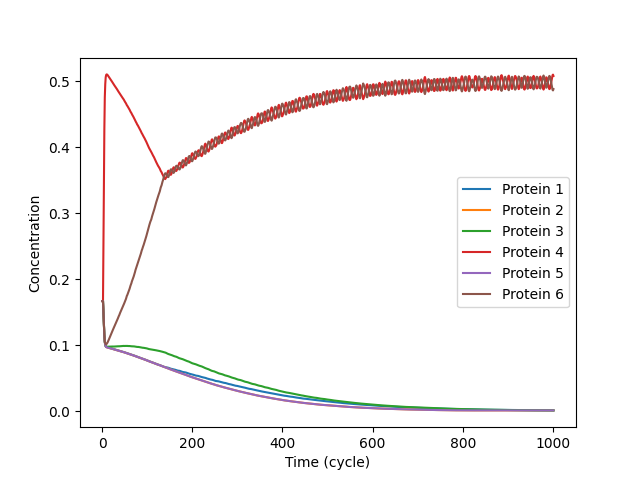} 
    \caption{Hybrid} 
    \label{dyna:c} 
  \end{subfigure}
   \begin{subfigure}[b]{0.45\linewidth}
    \centering
    \includegraphics[width=0.95\linewidth]{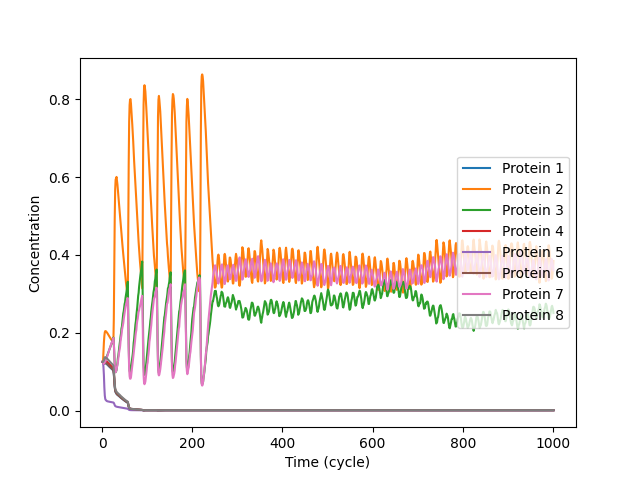} 
    \caption{Chaotic} 
    \label{dyna:d} 
  \end{subfigure}
  \caption{Different types of observed dynamics made by the proposed \ac{ARN}}
  
  \label{fig:dyna} 
\end{figure}

Figure \ref{fig:dyna} illustrates generated protein concentration dynamics belonging to four classes of patterns. In some cases protein concentrations develop over time following simple patterns (Figure \ref{dyna:a}). A shared characteristic of such dynamics is the smooth development of protein concentrations to reach a stable state where the concentrations do not vary over time. Figure \ref{dyna:b} shows an oscillatory dynamic in which one or more proteins produce a repeating pattern of concentration levels. This is a common behavior in protein dynamics and is often accomplished by two \ac{TF} types competing to achieve higher production levels. Figure \ref{dyna:c} illustrates a Hybrid behavior in which both oscillation (concentration levels of proteins 4 and 6) and simple development can be observed. Finally, Figure \ref{dyna:d} shows a more chaotic dynamic behavior. Until around cycle 230, the network produces an oscillatory dynamic between proteins 3, 2, and 7 which seems to be stabilizing; however, the dynamics change to a different type of oscillation after this cycle with ostensibly unique intervals. 

\begin{figure}[ht] 
  \begin{subfigure}[b]{0.45\linewidth}
    \centering
    \includegraphics[width=0.95\linewidth]{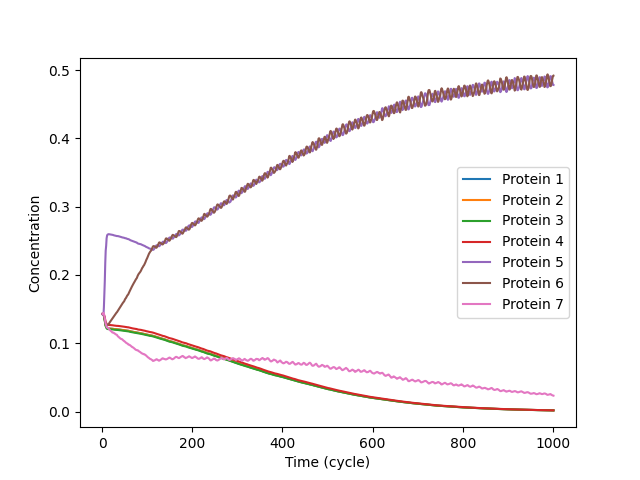} 
    \caption{Protein Dynamic} 
    \label{rate:a} 
  \end{subfigure}
  \begin{subfigure}[b]{0.45\linewidth}
    \centering
    \includegraphics[width=0.95\linewidth]{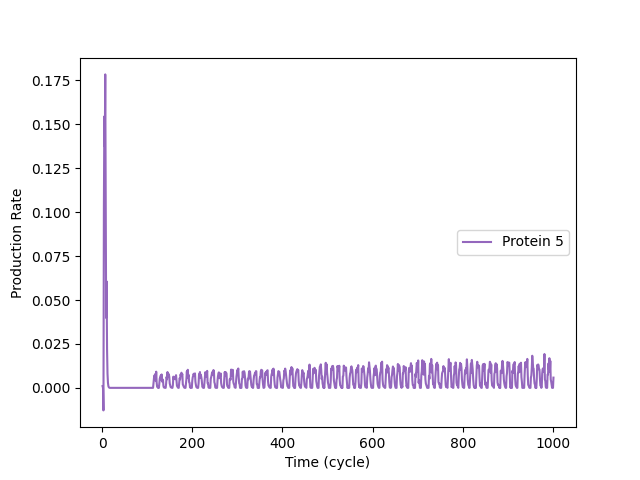} 
    \caption{Production Signals Over Time} 
    \label{rate:b} 
  \end{subfigure} 
  
  \caption{Dynamics vs. Transcription Rate. a) A Hybrid produced protein dynamic b) Transcription rates produced for Protein 5 over time}
  
  \label{fig:rate} 
\end{figure}

Figure \ref{fig:rate} illustrates a protein dynamic (Figure \ref{rate:a}) and the production rates/signals overtime responsible for regulating these dynamics of protein 5 (Figure \ref{rate:b}). The production rate increases if more and stronger \ac{TF} binding happens on the enhancer region of the gene compared to the inhibitor region. If no binding happens, the production rate will be zero. At the start of the regulatory cycles, \ac{TF}s have to move in the cell to reach the regulatory sites of genes other than their parents. The first few bonds result in a more intense increase/decrease in production rates since other \ac{TF}s need to also move and spread in the cell to start stabilizing the network of the interaction (Increase in the production level of Protein 5 in Figure \ref{rate:b} during the first 10 cycles). In Figure \ref{rate:b}, this is followed by a steady no-production state for 100 cycles in which a drop in protein concentrations can be noticed in the protein dynamics. No production signal or rates less than 0 can be considered equivalents of natural genes not being expressed or turned off by natural repressors. After cycle 150, an oscillatory behavior for Protein 5 can be observed in the dynamics. This behavior is correlated with the oscillatory patterns of production rates.    

\subsection{Impact of Initial States on the System Dynamics}

In this section, the impact of the initial state and parameters of the system on the outcome of the protein dynamics is studied. The random state for the movement of \ac{TF}s is preserved in all cases, so the stochastic nature of the proposed system does not make the comparisons unfair, and only a single parameter changes for each comparison.

\textbf{Genome Size vs. Number of Genes:} Table \ref{table:genecount} demonstrates the average number of genes that were found following the promoter-terminator rule in a genome with different sizes for 100 individuals. It can be deduced that the number of genes almost linearly increases with the genome length. On average, one to two new genes is found for every 1000 lengths added to the genome.

\begin{table}[ht]
\resizebox{\textwidth}{!}{%
\begin{tabular}{|
>{\columncolor[HTML]{EFEFEF}}l |l|l|l|l|l|l|l|l|l|l|}
\hline
\textbf{Genome Length} & 1000 & 2000 & 3000 & 4000 & 5000 & 6000 & 7000 & 8000 & 9000 & 10000 \\ \hline
\textbf{Genes} & 2 & 4 & 6 & 7 & 9 & 11 & 13 & 15 & 17 & 19 \\ \hline
\end{tabular}%
}
\caption{Average count of genes witnessed for different DNA sizes among 100 individual runs. Gene numbers are rounded values.}
\label{table:genecount}
\end{table}

\begin{figure}[H] 
  \begin{subfigure}[b]{0.45\linewidth}
    \centering
    \includegraphics[width=0.95\linewidth]{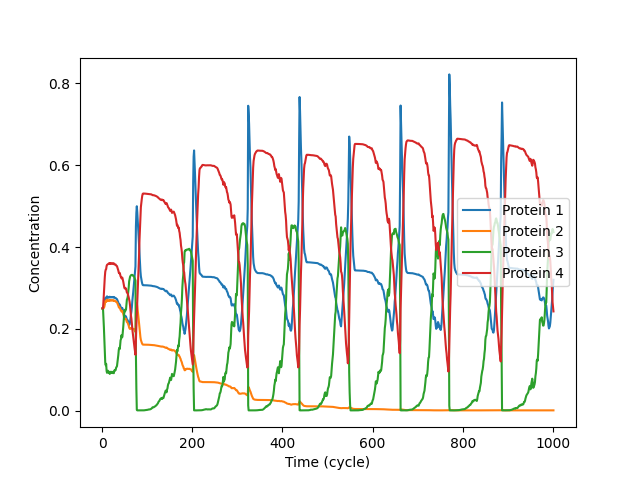} 
    \caption{Original Dynamic} 
    \label{comp:a} 
  \end{subfigure}
  \begin{subfigure}[b]{0.45\linewidth}
    \centering
    \includegraphics[width=0.95\linewidth]{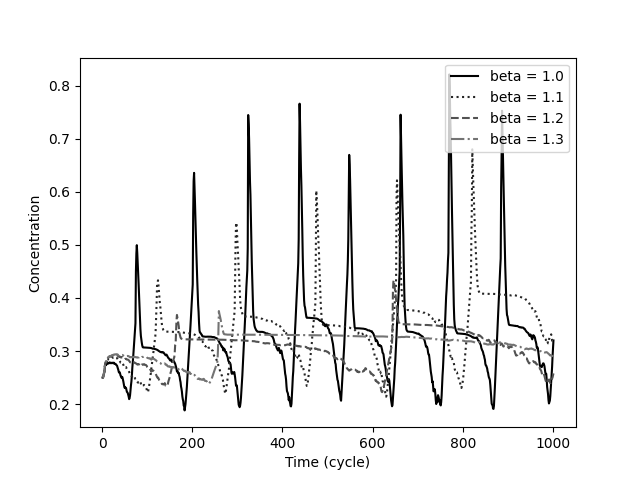} 
    \caption{$\beta$} 
    \label{comp:b} 
  \end{subfigure} 
  \begin{subfigure}[b]{0.45\linewidth}
    \centering
    \includegraphics[width=0.95\linewidth]{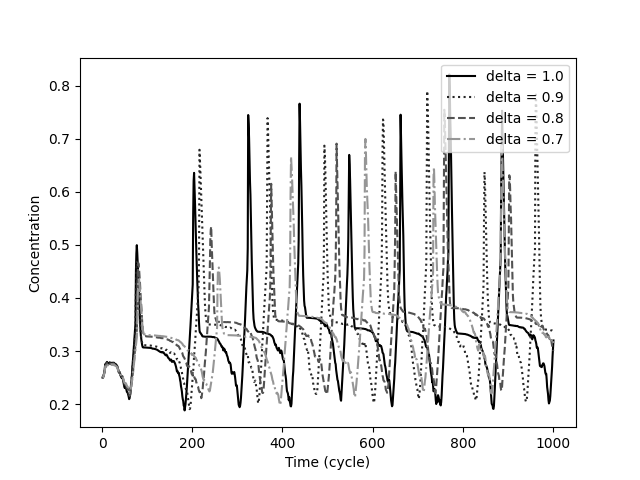} 
    \caption{$\delta$} 
    \label{comp:c} 
  \end{subfigure}
   \begin{subfigure}[b]{0.45\linewidth}
    \centering
    \includegraphics[width=0.95\linewidth]{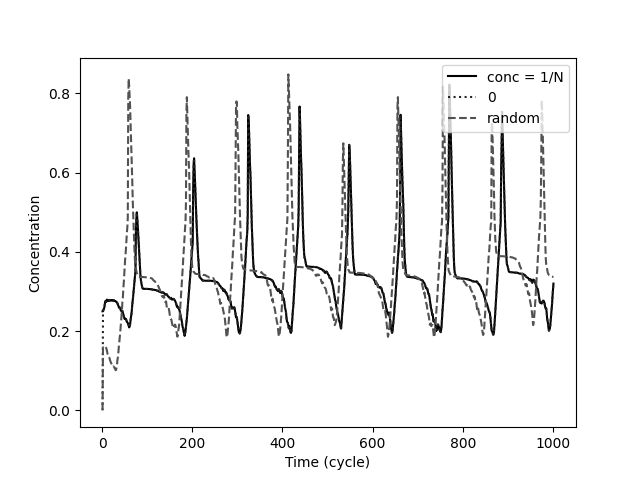} 
    \caption{Starting Concentrations} 
    \label{comp:d} 
  \end{subfigure}
  \\
  \begin{subfigure}[b]{0.45\linewidth}
    \centering
    \includegraphics[width=0.95\linewidth]{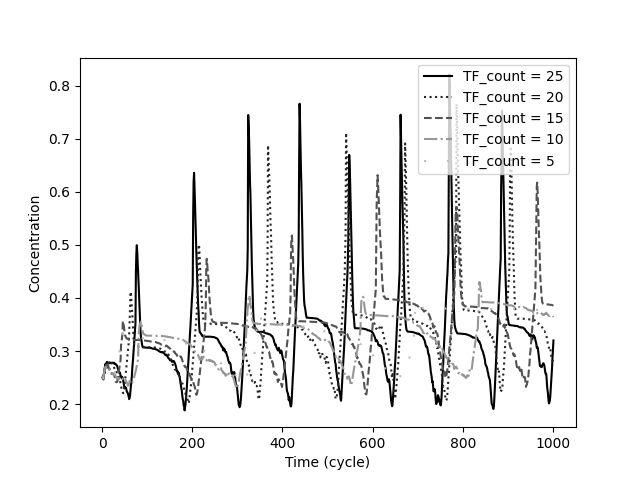} 
    \caption{Transcription Factors} 
    \label{comp:e} 
  \end{subfigure}
   \begin{subfigure}[b]{0.45\linewidth}
    \centering
    \includegraphics[width=0.95\linewidth]{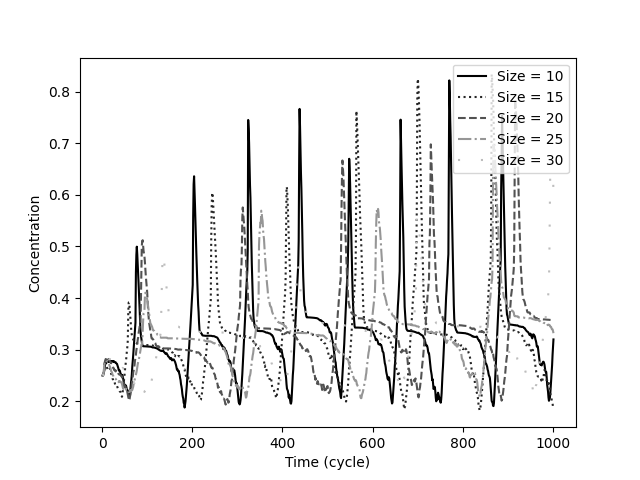} 
    \caption{Cell Size} 
    \label{comp:f} 
  \end{subfigure}
  \caption{An study of the impact of different initial conditions on the produced dynamics by the system}
  
  \label{fig:compare} 
\end{figure}

\textbf{Arbitrary Constant $\beta$ :} $\beta$ is an arbitrary constant that can be served to control the intensity of the inhibitory and enhancing signals. The value of $\beta$ is initially set to 1 for most experiments. Figure \ref{comp:a} shows a protein dynamic selected as a base dynamic to compare the impact of different initial states. Figure \ref{comp:b} illustrates how different $\beta$ values changes the dynamics for Protein 1. With an increase in the $\beta$ value, a shift in the generated patterns can be seen in time in a way that the same patterns happen later in the regulatory cycles but on a smaller scale. For $\beta = 1.3$, the development is so slow that only one spike pattern can be observed during 1000 cycles compared to the eight spike patterns of $\beta = 1$.

\textbf{$\delta$ Arbitrary Constant :} This constant value controls the intensity of protein production and is multiplied by the production signal at each development cycle. Same as $\beta$, the initial value of $\delta$ is set to 1. Figure \ref{comp:c} compares different $\delta$ values and their impact on the generated dynamics. A shift in time for the generated patterns can be seen for less $\delta$ values. In other words, lower $\delta$ values expand the produced dynamics while higher $\delta$ values shrink it. This impact is the opposite of the impact of $\beta$ on the dynamics. Also, unlike $\beta$, the concentration levels seem not to change as much, and the scale of the dynamics remains closer to the original one. 

\textbf{Initial Protein Concentration Levels:} The initial concentration levels of the proteins were set to $\frac{1}{N}$ ($N$ is the number of genes) for most of our experiments. Here, an experiment was conducted to see the impact of these initial conditions by trying 0 and random initial levels. For the case that initial concentrations are set to 0 (Figure \ref{comp:d}), no change in the produced dynamics can be observed. Although not included in the graph, different constant values for initial concentrations were attempted. However, in all cases where the initial concentration levels were similar for all the proteins, no change in dynamics could be observed. Setting initial concentration levels to a random value resulted in a similar dynamic with the same patterns slightly shifted in different directions. 

\textbf{Starting TF Count:} Another interesting factor to study is the impact of the initial number of \ac{TF}s for each gene on the produced dynamics. It can be observed in Figure \ref{comp:e} that a change in the \ac{TF} counts causes a shift in time and scale of the produced dynamics. It is worth noting that, unlike the case for $\beta$ and $\delta$ the time shift seems random and happens in both directions. An early appearance of the spike pattern happens in approximately cycle 60, where as the pattern occurs in a latter stage around approximately cycle 220. 

\textbf{Cell/Grid Size:} The same as the case for different \ac{TF} values, a shift in time and scale can be seen for different cell sizes (Figure \ref{comp:f}). The randomness in the scale and time shift is due to the random movements of the \ac{TF}s. The larger the cell is, the longer it takes for the \ac{TF}s to spread in the grid and to help the network to stabilize. 

\begin{figure}[ht]
    \centering
    \includegraphics[width=0.9\columnwidth]{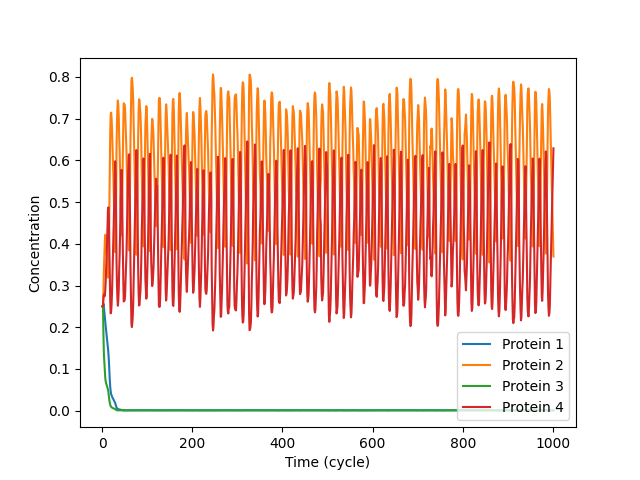}
    \caption{Dynamics produced after a slight change in the position of the enhancer site of gene 1 of the network that produced Figure \ref{comp:a} dynamics}
    \label{fig:enhpos}
\end{figure}

\textbf{Changing Spatial Position of the Regulatory Sites:} The proposed system is robust. Changing most initial states of the system, do not alter the shape of the generated patterns, but alters their scale or shifts them in time. However, a slight change in the position of the regulatory sites in the 2D grid, might significantly change the dynamics produced. In Figure \ref{fig:enhpos} the enhancer position of gene 1 of the same network from Figure \ref{comp:a} is moved one unit away from its original position. A different protein concentration dynamic is generated in which the concentration level of proteins 1 and 3 stabilize to a value close to 0 in early cycles and concentration levels of proteins 2 and 4 produce an oscillatory pattern. Similar experiment was made on the other regulatory sites of different networks. In most cases, the dynamics drastically change; however, there are cases in which the dynamics slightly shift or do not change at all.

\subsection{Evolution of Dynamics}

So far, the nature of the proposed system has been explained, and the different dynamics produced from random genomes generated from random seeds have been studied. However, to apply the proposed \ac{ARN} in other applications, it is essential for this system to be evolvable to achieve desired dynamics. In this section, two experiments were conducted to evolve regulatory networks that meet a specific dynamic criterion. To evolve these networks, a simple \ac{GA} was used that alters the initial DNA genotype of each individual. The utilized \ac{GA} consists of a population of genome, one-point crossover, point mutation, and tournament as the selection mechanism. For both evolutionary experiments, a population size of 25, a mutation rate of 0.10, and a tournament size of 3 were configured. Each experiment was run for 50 generations.

\begin{figure}[ht] 
  \begin{subfigure}[b]{0.45\linewidth}
    \centering
    \includegraphics[width=0.95\linewidth]{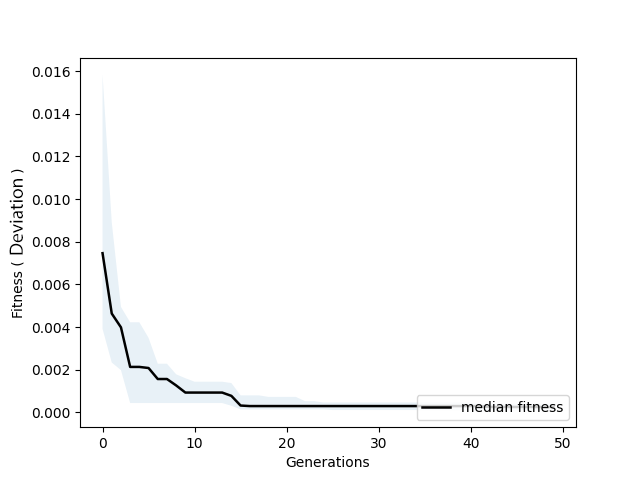} 
    \caption{Evolution of Individuals for Problem 1} 
    \label{evo:a} 
  \end{subfigure}
  \begin{subfigure}[b]{0.45\linewidth}
    \centering
    \includegraphics[width=0.95\linewidth]{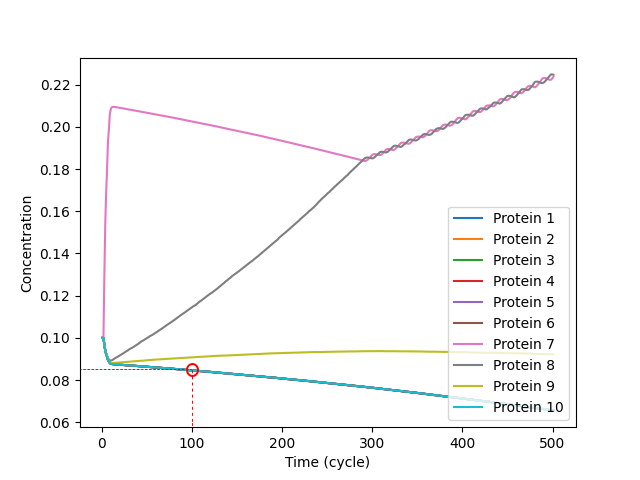} 
    \caption{Problem 1 Solution} 
    \label{evo:b} 
  \end{subfigure} 
  \begin{subfigure}[b]{0.45\linewidth}
    \centering
    \includegraphics[width=0.95\linewidth]{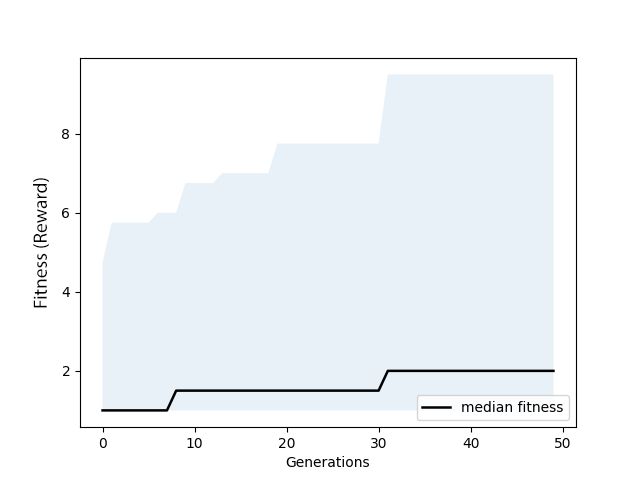} 
    \caption{Evolution of Individuals for Problem 2} 
    \label{evo:c} 
  \end{subfigure}
   \begin{subfigure}[b]{0.45\linewidth}
    \centering
    \includegraphics[width=0.95\linewidth]{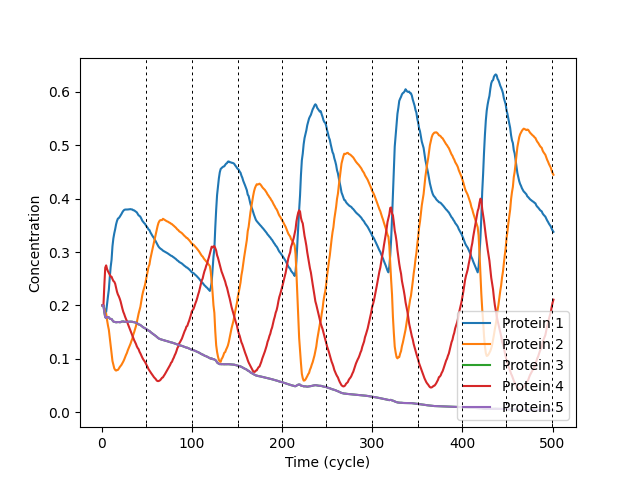} 
    \caption{Problem 2 Solution} 
    \label{evo:d} 
  \end{subfigure}
  \caption{Evolution of dynamics}
  
  \label{fig:evo} 
\end{figure}

In problem 1, the goal is for Protein 1 to reach a 0.085 concentration level at cycle 100. Figure \ref{evo:a} shows the evolutionary results for this problem. The X-axis is time, and the Y-axis is the deviation from the goal concentration in the form of absolute error. Therefore, lower values indicate a better individual. The depicted line represents the median fitness for the experiment's 10 parallel runs, and the shaded areas represent the 75 and 25 quantiles. Figure \ref{evo:b} illustrates one of the evolved solutions for solving this problem. Several proteins share the same concentration levels. 

In problem 2, the goal is for proteins 1 and 2 to alternate in concentration levels every 50 cycles so that in the starting period if protein 1 has more concentration than protein 2, the individual will be rewarded by 1 point. The individual receives another reward if, in the next period, protein 2 has more concentration than 1. The same level alteration process should continue for 10 cycle periods to achieve the maximum reward of 10. This is a more challenging task than problem 1, and the considered fitness function based on discrete rewards does not provide significant pressure towards solving the problem. Figure \ref{evo:c} shows the evolutionary results for solving problem 2. Although the median of individuals does not solve the problem, some cases fully solve it in 50 generations. Figure \ref{evo:d} shows a perfect solution to the problem achieved during evolution by one of the runs with fitness equal to 10.

\begin{figure}[ht]
    \centering
    \includegraphics[width=0.9\columnwidth]{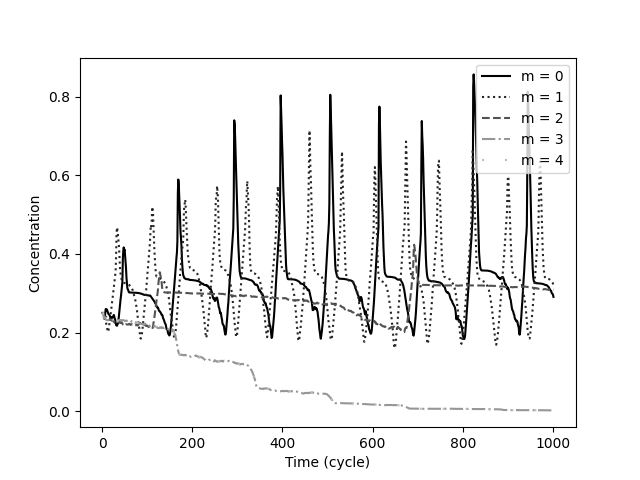}
    \caption{A comparison between the impact of different number of mutations on expressed genes}
    \label{fig:mutation}
\end{figure}

Lastly, we conduct an experiment in which an attempt is made to depict the impact of point mutations on the outcome of the dynamics. Mutations occurring on different sites result in different behaviors, and their impact highly depends on the rest of the expressed genome. A single point-mutation on the protein-coding site can sometimes completely change the system's dynamics, while sometimes it serves as a neutral mutation. In general, during the experiments, three significant outcomes from mutations could be observed in the system: 1- a neutral mutation, 2- a complete change in the dynamics, and 3- a shift in the scale and time of the dynamics. Figure \ref{fig:mutation} shows the concentration levels of only protein 1 from the entire dynamic of \ref{comp:a} undergoing 0 to 5 point mutations on the regulatory sites of the expressed genes. In the case of only one mutation, the concentration dynamics shrink, and more of the same spike patterns can be seen in the same number of cycles. In the case of two mutations, the dynamic expands, and only two of the spike patterns occur. The case of 3 mutations completely changes the dynamics, although the fourth mutation does not change the dynamics any further. It can be seen that more mutations have a higher probability of changing the dynamics entirely.

\section{Discussion and Conclusion}
\label{discussionsec}

In this paper, a biologically-close \ac{ARN} model was introduced based on the work of \cite{Banzhaf2003} in which \ac{TF}s control the regulation of protein productions to produce various protein dynamics. A \ac{CA} was utilized to introduce the spatial properties to the system. The rules of interactions between proteins and regulatory regions were defined by an \ac{AC}. Our results showed protein dynamics produced, close to the biological counterparts, and a classification of these dynamics was performed. The impact of the initial states of the system on the produced dynamics and how they can help control the outcome of the system were explored. An interesting take on these experiments was the controllable heterochrony in the proposed system. These results indicated that the proposed system is highly robust and changing most of the initial states of the system do not change the dynamics produced. However, a slight change in the spatial position of the regulatory sites on the 2D grid can drastically change these dynamics which can be a means for providing inputs to the system. The proposed networks were evolved using a simple evolutionary algorithm to solve two problems that specify the states of the produced dynamics at specific cycle periods. Finally, the impact of the mutation on the produced dynamics was studied, which showed the high evolvability of the proposed system. In future, employing techniques such as Dynamic Time Warping and Compression-based Dissimilarity Measures could be helpful to analyze and differentiate the produced dynamics systematically. \ac{ARN} representations were previously used as direct and indirect representations for genetic programming. It would be worthwhile to try the proposed representation for genetic programming to solve computational problems.    

\bibliographystyle{ieeetr} 
\bibliography{bib}

\end{document}